\documentclass[p2]{elsarticle}

\usepackage{hyperref}

\journal{Journal of Neurocomputing SI-DL-ISR19 (Accepted)}

\usepackage[normalem]{ ulem }
\usepackage{soul}
\usepackage{color}
\usepackage[table]{xcolor}
\usepackage{amsfonts}       
\usepackage{amsmath}
\usepackage{bbding}
\usepackage{pifont}
\usepackage{wasysym}
\newcommand{\cmark}{\ding{51}}%
\newcommand{\xmark}{\ding{55}}%
\usepackage{array,multirow,graphicx}
\usepackage{float}
\usepackage{mwe}    
\usepackage{subfig}
\usepackage{adjustbox}
\usepackage{comment}
\DeclareMathOperator*{\argmin}{arg\,min}

\DeclareMathOperator*{\argmax}{arg\,max}
\DeclareMathOperator*{\Id}{Id}

\begin{document}

\begin{frontmatter}

\title{Generative Collaborative Networks\\ for Single Image Super-Resolution}


\author[mymainaddress,mysecondaryaddress]{Mohamed El Amine Seddik}
\ead{mohamedelamine.seddik@cea.fr}

\author[mymainaddress]{Mohamed Tamaazousti}
\ead{mohamed.tamaazousti@cea.fr}

\author[mymainaddress,johnaddress]{John Lin}
\ead{john.lin@cea.fr}

\address[mymainaddress]{CEA/LIST/DIASI/LVIC, F-91191 Gif-sur-Yvette, France}
\address[mysecondaryaddress]{CentraleSupelec/L2S, 3 rue Joliot Curie, 91192 Gif-sur-Yvette, France}
\address[johnaddress]{ISIT-UMR6284 CNRS/Auvergne University, Clermont-Ferrand, France}

\begin{abstract}
A common issue of deep neural networks-based methods for the problem of Single Image Super-Resolution (SISR), is the recovery of finer texture details when super-resolving at large upscaling factors. This issue is particularly related to the choice of the objective loss function. In particular, recent works proposed the use of a VGG loss which consists in minimizing the error between the generated high resolution images and ground-truth in the feature space of a Convolutional Neural Network (VGG19), pre-trained on the very ``large'' ImageNet dataset. When considering the problem of super-resolving images with a distribution ``far'' from the ImageNet images distribution (\textit{e.g.,} satellite images), their proposed \textit{fixed} VGG loss is no longer relevant. In this paper, we present a general framework named \textit{Generative Collaborative Networks} (GCN), where the idea consists in optimizing the \textit{generator} (the mapping of interest) in the feature space of a \textit{features extractor} network. The two networks (generator and extractor) are \textit{collaborative} in the sense that the latter ``helps'' the former, by constructing discriminative and relevant features (not necessarily \textit{fixed} and possibly learned \textit{mutually} with the generator). We evaluate the GCN framework in the context of SISR, and we show that it results in a method that is adapted to super-resolution domains that are ``far'' from the ImageNet domain.
\end{abstract}


\end{frontmatter}


\begin{figure}
\centering
\includegraphics[width=\linewidth]{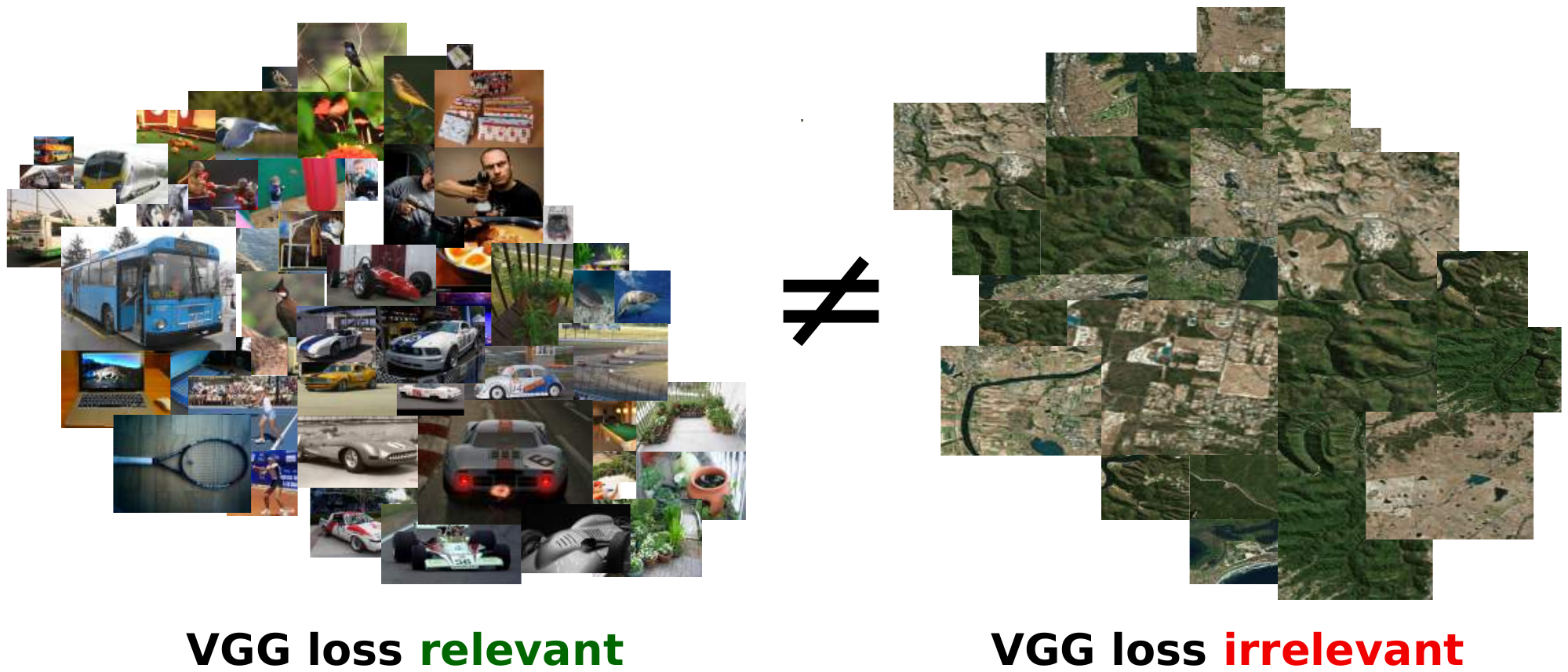}
\caption{When super-resolving images from a different domain (\textit{e.g.}, satellite images on the right) than the ImageNet domain (\textit{e.g.}, general objects on the left), the VGG loss introduced by \citep{ledig2016photo} is no longer relevant. We propose a method that outperforms the SRGAN method \citep{ledig2016photo} when super-resolving satellite images. Our method falls within a large class of methods which constitutes our proposed \textit{Generative Collaborative Networks} framework.}
\end{figure}

\section{Introduction}
The super-resolution problem ($\mathcal{P}_{sr}$) consists in estimating a high resolution (HR) image from its corresponding low resolution (LR) counterpart.
$\mathcal{P}_{sr}$ finds a wide range of applications and has attracted much attention within the community of computer vision \citep{nasrollahi2014super, yang2007spatial,zou2012very}. Generally, the considered optimization objective of supervised methods to solve $\mathcal{P}_{sr}$ is the minimization of the mean squared error (MSE) between the recovered HR image and ground-truth. This class of methods are known to be suboptimal to reconstruct texture details at large upscaling factors. In fact, since MSE consists in a pixel-wise images differences, its ability to recover high texture details is limited \cite{ledig2016photo,gupta2011modified,wang2004image, wang2003multiscale}. Furthermore, the minimization of MSE maximizes the Peak Signal-to-Noise-Ratio (PSNR) metric, which is commonly used for the evaluation of $\mathcal{P}_{sr}$ methods \citep{yang2014single}.\par
In order to correctly recover finer texture details when super-resolving at large upscaling factors, a recent (state-of-the-art) work \citep{ledig2016photo} defined a perceptual loss which is a combination of an adversarial loss and a VGG loss. The former encourages solutions perceptually hard to distinguish from the HR ground-truth images, while the latter consists in using high-level feature maps of the VGG network \citep{simonyan2014very} pre-trained on ImageNet \cite{deng2009imagenet}. When considering the problem of super-resolving images from a target-domain \textit{different} than ImageNet (\textit{e.g.,} satellite images), the features produced by the pre-trained VGG network on the source domain (ImageNet) are suboptimal and no longer relevant for the target domain. In fact, transfer-learning methods are known to be efficient only when the source and target domains are close enough \cite{tamaazousti2017mucale,tamaazousti2018universal,karbalayghareh2018optimal}. In this work, we present a general framework which we call \textit{Generative Collaborative Networks} (GCN), where the main idea consists in optimizing the generator (\textit{i.e.}, the mapping of interest) in the feature space of a network which we shall refer to as a \textit{features extractor} network. The two networks are said to be \textit{collaborative} in the sense that the features extractor network ``helps'' the generator by constructing (here, learning) relevant features. In particular, we applied our framework to the problem of single image super-resolution, and we demonstrated that it results in a method that is more adapted (compared to SRGAN \citep{ledig2016photo}) when super-resolving images from a domain that is ``far'' from the ImageNet domain.\par 
The rest of the paper is organized as follows. In Section~\ref{sec:sofa} we present the state of the art on the problem of single image super-resolution. We describe our Generative Collaborative Networks framework in Section~\ref{sec:GCN}. Section~\ref{sec:application} presents our proposed method for the super resolution task and related experimental results. Section~\ref{sec:disc} provides some discussions and concludes the article.

\section{Related work} 
\label{sec:sofa}
The problem of super-resolution has been tackled with a large range of approaches. In the following, we will consider the problem of \textit{single} image super-resolution ($\mathcal{P}_{sisr}$) and thus the approaches that recover HR images from multiple images \citep{borman1998super, farsiu2004fast} are out of the scope of this paper. First approaches to solve $\mathcal{P}_{sisr}$ were filtering-based methods (\textit{e.g.}, linear, bicubic or Lanczos \citep{duchon1979lanczos} filtering). Even if these methods are generally very fast, they usually yield overly smooth textures solutions \cite{wang2004image}. Most promising and powerful approaches are learning-based methods which consist in establishing a mapping between LR images and their HR counterparts (supposed to be known). Initial work was proposed by Freeman \textit{et al.} \citep{freeman2002example}. This method has been improved in \citep{dong2011image, zeyde2010single} by using compressed sensing approaches. Patch-based methods combined with machine learning algorithms were also proposed: in \citep{timofte2013anchored, timofte2014a+} upsampling a LR image by finding similar LR training patches in a low dimensional space (using neighborhood embedding approaches) and a combination of the HR patches counterparts are used to reconstruct HR patches. A more general mapping of example pairs (using kernel ridge regression) was formulated by Kim and Kwon \citep{kim2010single}. Similar approaches used Gaussian process regression \citep{he2011single}, trees \citep{salvador2015naive} or Random Forests \citep{schulter2015fast} to solve the regression problem introduced in \citep{kim2010single}. An ensemble method-based approach was adopted in \citep{dai2015jointly} by learning multiple patch regressors and selecting the most relevant ones during the test phase.\par
Convolutional neural networks (CNN)-based approaches outperformed other $\mathcal{P}_{sisr}$ approaches, by showing excellent performance. Authors in \citep{wang2015deep} used an encoded sparse representation as a prior in a feed-forward CNN, based on the learned iterative shrinkage and thresholding algorithm   of \citep{gregor2010learning}. An end-to-end trained three layer deep fully convolutional network, based on bicubic interpolation to upscale the input images, was used in \citep{dong2014learning, dong2016image} and achieved good $\mathcal{P}_{sisr}$ performances. Further works suggested that enabling the network to directly learn the upscaling filters, can remarkably increase performance in terms of both time complexity and accuracy \citep{dong2016accelerating, shi2016real}. In order to recover visually more convincing HR images, Johnson \textit{et al.} \citep{johnson2016perceptual} and Bruna \textit{et al.} \citep{bruna2015super} used a closer loss function to perceptual similarity. More recently, authors in \citep{ledig2016photo} defined a perceptual loss which is a combination of an adversarial loss and a VGG loss. The latter consists in minimizing the error between the recovered HR image and ground-truth in the high-level feature space of the pre-trained VGG network \citep{simonyan2014very} on ImageNet \cite{deng2009imagenet}. This method notably outperformed CNN-based methods for the problem $\mathcal{P}_{sisr}$.

\section{Generative Collaborative Networks}
\label{sec:GCN}
\subsection{Proposed Framework}
Consider a problem $\mathcal{P}$ of learning a mapping function $\mathcal{F}$, parameterized by $\theta_{\mathcal{F}}$, that transforms images from a domain $\mathcal{X}$ to a domain $\mathcal{Y}$, given a training set of $N$ pairs $\{ (x_i,y_i)\}_{i=1}^N\in \mathcal{X}\times \mathcal{Y}$. Denote by $p_{\mathcal{X}}$ and $p_{\mathcal{Y}}$ the probability distributions respectively over $\mathcal{X}$ and $\mathcal{Y}$. In addition, we introduce a given \textit{features extractor} function denoted $\Phi$, parameterized by $\theta_{\Phi}$, that maps an image $y\in \mathcal{Y}$ to a certain euclidean feature space $\mathcal{S}_{\Phi}$ of dimensionality $d$. The mappings $\mathcal{F}$ and $\Phi$ are typically feed-forward Convolutional Neural Networks. The Generative Collaborative Networks (GCN) framework consists in learning the mapping function $\mathcal{F}$ by minimizing a given loss function\footnote{$\ell_2$-loss is considered in the following.} in the space of features $\mathcal{S}_{\Phi}$, between the generated images (through $\mathcal{F}$) and ground-truth. Formally,
\begin{align}
\hat{\theta}_{\mathcal{F}} = \argmin_{\theta_{\mathcal{F}}} \frac{\lambda_1}{N\, d} \sum_{i=1}^N \sum_{j=1}^d \left( \Phi_j \left( y_i \right) - \Phi_j \left( \mathcal{F}(x_i) \right)  \right)^2 + \lambda_2\, \Omega(\theta_{\mathcal{F}}),
\label{eq:obj}
\end{align}
where $\Omega(\theta_{\mathcal{F}})$ is a certain regularization term (detailed below) on the weights $\theta_{\mathcal{F}}$ and $\lambda_1$ and $\lambda_2$ are summation coefficients. The two networks $\mathcal{F}$ and $\Phi$ are collaborative in the sense that, the latter learns specific features of the domain $\mathcal{Y}$ and ``helps'' the former, as it is learned in the space $\mathcal{S}_{\Phi}$. An important question arises about how to learn the mapping $\Phi$. In following, we describe different classes of methods depending on the learning strategy of $\Phi$. In fact, the features extractor function $\Phi$ can take different forms and be learned by different strategies. In particular, we distinguish two learning strategies (illustrated in Figure~\ref{fig:overview}), which we shall call \textit{disjoint-learning} and \textit{joint-learning}. The four following cases belong to the \textit{disjoint-learning} strategy:
\begin{figure}[h!]
\centering
\includegraphics[width=0.95\textwidth]{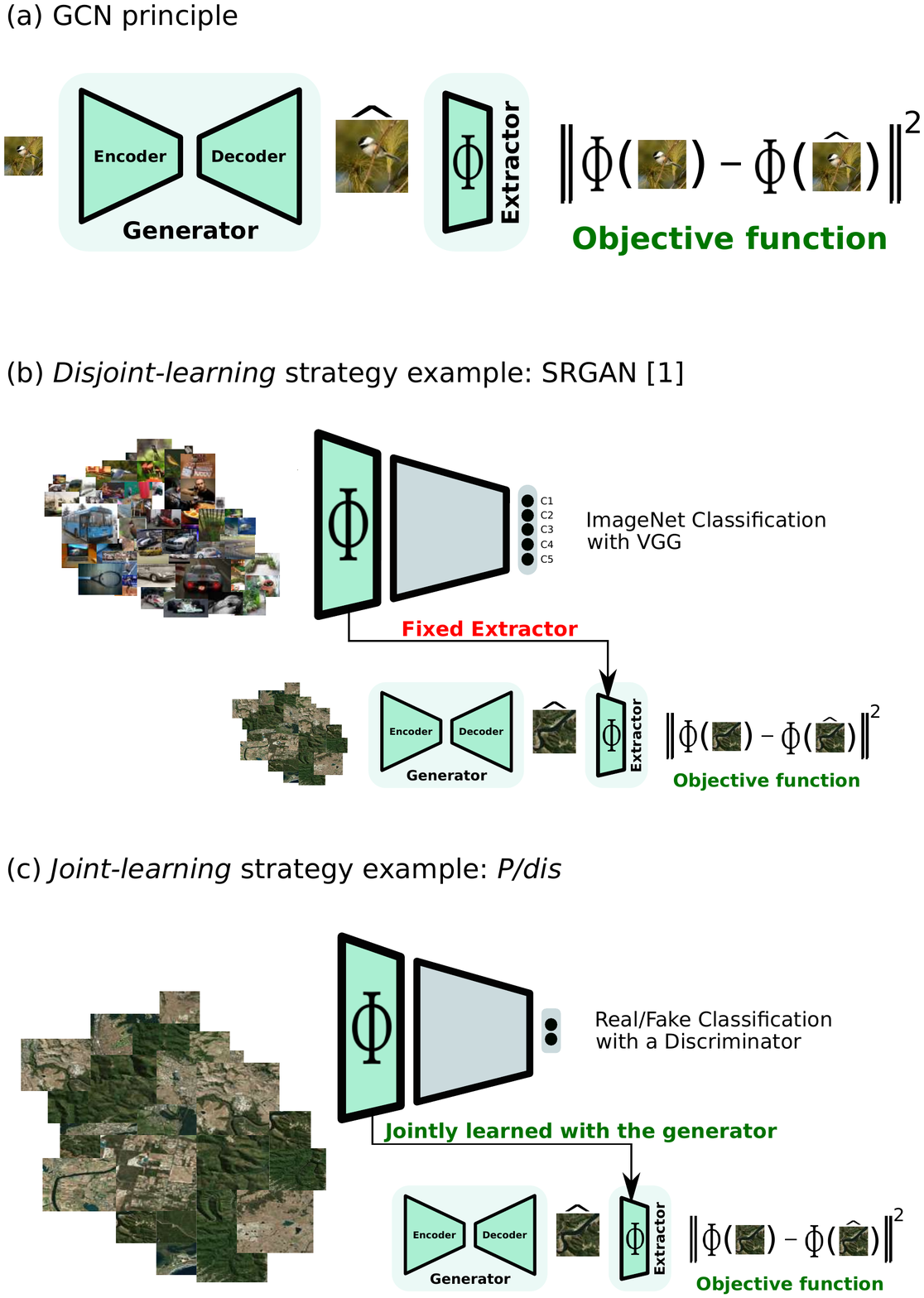}
\caption{Overview of the GCN framework with examples of the two learning strategies. The GCN framework consists in optimizing a \textit{generator} in the feature space of an \textit{extractor} as illustrated in \textbf{(a)}. The extractor can be trained beforehand and used to optimize the generator, which we refer to as \textit{disjoint-learning} strategy \textbf{(b)}. The extractor can also be optimized jointly with the generator, i.e., using a \textit{joint-learning} strategy \textbf{(c)}.}
\label{fig:overview}
\end{figure}
\begin{itemize}
\item[(1.a)] When $\Phi$ is the \textit{identity operator} ($\Phi = \Id$). In that case, the objective in Eq.\eqref{eq:obj} becomes a simple pixel-wise MSE loss function. We refer to this class of methods by $\mathcal{P}$/\textit{mse}.
\item[(1.b)] When $\Phi$ corresponds to a \textit{random feature map} neural network, that is to say, the weights $\theta_{\Phi}$ are set randomly according to a given distribution $\mu$. We refer to this class of methods by $\mathcal{P}$/\textit{ran}.
\item[(1.c)] When $\Phi$ is a part of a model that solves a \textit{reconstruction problem} (jointly with an auxiliary mapping function $\Psi:\mathcal{S}_{\Phi}\rightarrow \mathcal{Y}$), by minimizing the pixel-wise $\ell_2$-loss function between the reconstructed images (through $\Psi$) and ground-truth:
\begin{align}
(\hat{\theta}_{\Phi},\_)= \argmin_{(\theta_{\Phi}, \theta_{\Psi})} \frac{1}{N\, \dim(\mathcal{Y})} \sum_{i=1}^N \sum_{j=1}^{\dim(\mathcal{Y})} \left( (y_i)_j - (\Psi\circ \Phi (y_i))_j\right)^2.
\label{eq:recon}
\end{align}
Notably, this strategy allows for the learning of reconstruction features which are different from classification-based features. We refer to this class of methods by $\mathcal{P}$/\textit{rec}.
\item[(1.d)] When $\Phi$ is trained to solve a \textit{multi-label classification problem} \cite{ledig2016photo}, that is to say, when labels are available for the domain $\mathcal{Y}$. More precisely, it exists a dataset $\left\lbrace (y_i, c_i) \right\rbrace_{i=1}^n\in \mathcal{Y}\times \{1,\ldots,m\}$ of $n$ images labelled among $m$ classes and $\Phi$ is learned to minimize the following objective:
\begin{align}
(\hat{\theta}_{\Phi},\_)= \argmax_{(\theta_{\Phi}, \theta_{\Psi})} \mathbb{P}\left\lbrace \Psi \circ \Phi(y_i) =c_i\, |\, y_i\, ;\, i\in \{1,\ldots,m\}\right\rbrace,
\end{align} 
where $\Psi:\mathcal{S}_{\Phi} \rightarrow \{1,\ldots,m\}$. We refer to this class of methods by $\mathcal{P}$/\textit{cla}.
\end{itemize}
The features extractor function $\Phi$ can also be trained \textit{jointly} with the desired mapping function $\mathcal{F}$. Indeed, as in the GANs paradigm, one can use a discriminator to distinguish the generated images (through $\mathcal{F}$) and ground-truth, and thus learn more relevant and specific features for the problem of interest $\mathcal{P}$. In particular, the \textit{joint-learning} strategy contains two cases:
\begin{itemize}
\item[(2.a)] When $\Phi$ is a part of a \textit{discriminator}. $\mathcal{D} = \Psi \circ \Phi:\mathcal{Y}\rightarrow\{0,1\}$ that classifies the generated images (through $\mathcal{F}$) and ground-truth. $\mathcal{D}$ is optimized in an alternating manner along with $\mathcal{F}$ to solve the adversarial min-max problem~\cite{sonderby2016amortised}:
\begin{align}
\min_{\theta_{\mathcal{F}}} \max_{  (\theta_{\Phi}, \theta_{\Psi })  } \mathbb{E}_{y\sim p_{\mathcal{Y}}} \left[ \log \Psi \circ \Phi (y) \right] + \mathbb{E}_{x\sim p_{\mathcal{X}}} \left[ \log \left\lbrace  1 - \Psi \circ \Phi\circ \mathcal{F}(x) \right\rbrace \right].
\label{eq:gan}
\end{align}
The adversarial loss (second term of Eq.~\eqref{eq:gan}) can thus be seen as a regularization of the parameters $\theta_{\mathcal{F}}$ by affecting this quantity to $\Omega(\theta_{\mathcal{F}})$ in Eq.~\eqref{eq:obj}. This regularization ``pushes'' the solution of the problem in Eq.~\eqref{eq:obj} to the manifold of the images in the domain $\mathcal{Y}$. We refer to this class of methods by $\mathcal{P}$/\textit{adv}. When $\lambda_2=0$, we refer to it by $\mathcal{P}$/\textit{dis}.
\item[(2.b)] When $\Phi$ is a part of a \textit{discriminator} and an \textit{auto-encoder}. Namely, by optimizing its weights $\theta_{\Phi}$ to solve simultaneously, an \textit{adversarial problem} as in Eq.~\eqref{eq:gan}; through $\mathcal{D} = \Psi_1 \circ \Phi:\mathcal{Y}\rightarrow\{0,1\}$, and a \textit{reconstruction problem} as in Eq.~\eqref{eq:recon}; through a mapping $\Psi_2:\mathcal{S}_{\Phi}\rightarrow \mathcal{Y}$. We refer to this class of methods by $\mathcal{P}$/\textit{adv,rec} or $\mathcal{P}$/\textit{dis,rec} depending on the value of $\lambda_2$ in Eq.~\eqref{eq:obj}.
\end{itemize}
\subsection{Existing Loss Functions}
The natural way to learn a mapping from a manifold to another is to use $\mathcal{P}$/\textit{mse} methods. It is well known \citep{ gupta2011modified, ledig2016photo, wang2003multiscale, wang2004image} that this class of methods lead to overly-smooth and poor perceptual quality solutions. In order to handle the mentioned perceptual quality limitation, a variety of methods have been proposed in the literature. First methods used generative adversarial networks (GANs) for generating high perceptual quality images \citep{denton2015deep, mathieu2015deep}, style transfer \citep{li2016combining} and inpainting \citep{yeh2016semantic}, namely the class of methods $\mathcal{P}$/\textit{adv} with $\lambda_1=0$. Authors in \citep{yu2016ultra} proposed to use $\mathcal{P}$/\textit{mse} with an adversarial loss ($\lambda_1>0$ and $\lambda_2>0$) to train a network that super-resolves face images with large upscaling factors. Authors in \citep{bruna2015super,johnson2016perceptual} and in \citep{dosovitskiy2016generating} used $\mathcal{P}$/\textit{cla} by considering respectively $\Phi=$VGG19 and $\Phi=$AlexNet networks as fixed features extractors (learned \textit{disjointly} from the mapping of interest), which result in a more perceptually convincing results for both super-resolution and artistic style-transfer \citep{gatys2015texture, gatys2016image}. More recently, authors in \citep{ledig2016photo} used $\mathcal{P}$/\textit{cla,adv} by considering $\Phi=$VGG19 as a fixed features extractor combined with an adversarial loss ($\lambda_2>0$). To the best of our knowledge, as summarized in table~\ref{tab:existent_loss}, the use of the other learning strategies of $\Phi$; namely (1.c), (2.a) and (2.b), have not been explored in the literature. We particularly apply these strategies in the context of Single Image Super-Resolution, which results in methods that are more suitable (comparing to the SRGAN method \citep{ledig2016photo}) to super-resolution domains that differ from the ImageNet domain. The proposed methods as well as the corresponding experiments are presented in the following section.  

\begin{table}[t!]
\scriptsize
\centering
\begin{tabular}{ c c c c c c }
Standard methods  & $\mathcal{P}/mse$ & $\mathcal{P}/cla$ & $\mathcal{P}/rec$ & $\mathcal{P}/dis$ & $\mathcal{P}/dis,rec$  \\ 
Existence  & \cmark \citep{gupta2011modified} & \cmark \citep{dosovitskiy2016generating} & \xmark & \xmark &  \xmark \\ 
\end{tabular}
\begin{tabular}{ c c c c c }
\hline
Adversarial methods  & $\mathcal{P}/adv,mse$ & $\mathcal{P}/adv,cla$ & $\mathcal{P}/adv$ & $\mathcal{P}/adv,rec$  \\ 
Existence  & \cmark \citep{yu2016ultra} & \cmark \cite{ledig2016photo} & \xmark & \xmark \\ 
\end{tabular}
\caption{Existent loss functions of the proposed GCN framework.}
\label{tab:existent_loss}
\end{table}

\section{Application of GCN to Single Image Super-Resolution}
\label{sec:application}
\subsection{Proposed Methods}
In this section, we consider the problem of Single Image Super-Resolution ($\mathcal{P}_{sisr}$). In particular, we suppose we are given $N$ pairs $\{ (I_i^{LR},I_i^{HR}) \}_{i=1}^N$ of low-resolution images and their high-resolution counterparts. Recalling our GCN framework (presented in Section~\ref{sec:GCN}) the proposed methods for the problem $\mathcal{P}_{sisr}$ are: $\mathcal{P}_{sisr}$/\textit{rec}, $\mathcal{P}_{sisr}$/\textit{dis}, $\mathcal{P}_{sisr}$/\textit{dis,rec}, $\mathcal{P}_{sisr}$/\textit{adv} and $\mathcal{P}_{sisr}$/\textit{adv,rec}. We show in the following that the most convincing results are given by $\mathcal{P}_{sisr}$/\textit{adv,rec}. In particular, we show on a dataset of satellite images (different from the ImageNet domain) that our method $\mathcal{P}_{sisr}$/\textit{adv,rec} outperforms the SRGAN method \citep{ledig2016photo} by a large margin on the considered domain. Note that, as our goal is to show the irrelevance of the VGG loss for some visual domains (different from ImageNet), we do not consider the well-known SR benchmarks (\textit{e.g.}, Set5, Set14, B100, Urban100) for the evaluation, as these benchmarks are relatively close to the ImageNet domain. The evaluation of the different methods is based on \textit{perceptual metrics} \citep{zhang2018unreasonable} which we recall in the following section.

\subsection{Evaluation Metrics}
The evaluation of super-resolution methods (more generally image regression-based methods) requires comparing visual patterns which remains an open problem in computer vision. In fact, classical metrics such as L2/PSNR, SSIM and FSIM often disagree with human judgments (\textit{e.g.}, blurring causes large perceptual change but small L2 change). Thus, the definition of a \textit{perceptual metric} which agrees with humans perception is an important aspect for the evaluation of $\mathcal{P}_{sisr}$ methods. Zhang \textit{et al.} \citep{zhang2018unreasonable} recently evaluated deep features across different architectures (Squeeze \cite{iandola2016squeezenet}, AlexNet\cite{krizhevsky2012imagenet} and VGG\cite{simonyan2014very}) and tasks (supervised, self-supervised and unsupervised networks) and compared the resulting metrics with traditional ones. They found that deep features outperform all classical metrics (\textit{e.g.}, L2/PSNR, SSIM and FSIM) by large margins on their introduced dataset. As a consequence, deep networks seem to provide an embedding of images which agrees surprisingly well with humans judgments.\par
Zhang \textit{et al.} \citep{zhang2018unreasonable} compute the distance between two images $x,y$ with a network\footnote{The considered networks are Squeeze\cite{iandola2016squeezenet}, AlexNet\cite{krizhevsky2012imagenet} and VGG\cite{simonyan2014very} and their "perceptual calibrated" versions which we refer to respectively as Squeeze-l, AlexNet-l and VGG-l. See \citep{zhang2018unreasonable} and the provided github project within for further details.} $\Phi$ in the following way: 
\begin{align}
d_{\Phi}(x,y) = \sum_l \frac{1}{H_l W_l} \sum_{h,w} \Vert w_l \odot (\Phi^l(x)_{hw} - \Phi^l(y)_{hw} ) \Vert_2^2,
\end{align}
where $\Phi^l(\cdot)$ are the extracted features from layer $l$ and unit-normalized in the channel dimension. $w_l$ is a re-scaling vector of the activations channel-wise at layer $l$. $H_l$ and $W_l$ are respectively the height and width of the $l^{th}$ feature map.\par
Thus, we compute the \textit{perceptual error} (PE) of a $\mathcal{P}_{sisr}$ method (a mapping $\mathcal{F}$) on a given test-set of $N$ low-resolution images and their high-resolution counterparts $\Pi=\{ (I_i^{LR},I_i^{HR}) \}_{i=1}^N$ as the mean distances between the generated images (through $\mathcal{F}$) and ground-truth as follows:
\begin{align}
\text{PE}_{\Phi}(\Pi) = \frac{1}{N}\sum_{i=1}^N d_{\Phi} (\mathcal{F}(I_i^{LR}), I_i^{HR}).
\label{eq:PE_metric}
\end{align}
Note that we use the implementation of \citep{zhang2018unreasonable} to compute the perceptual distances $d_{\Phi}(\cdot,\cdot)$ using six variants which are based on the networks Squeeze\cite{iandola2016squeezenet}, AlexNet\cite{krizhevsky2012imagenet} and VGG\cite{simonyan2014very} and their ``perceptual calibrated'' versions. The best method is considered to be the one which minimizes the maximum amount of PEs across different networks $\Phi\in \{\text{Squ, Squ-l, Alex, Alex-l, VGG, VGG-l}\}$.

\subsection{Experiments}
The overall goal of this section is to validate our statement about the relevance of the VGG loss when super-resolving images from a different domain than the ImageNet domain. To highlight this aspect, we first present the considered datasets, architectures and training details. Then we select the more appropriate method (across the GCN framework methods) for the $\mathcal{P}_{sisr}$ problem based on perceptual metrics \citep{zhang2018unreasonable}. Finally, we compare our proposed method to some baselines and the state-of-the-art SRGAN method \citep{ledig2016photo}, on three different datasets (detailed in the following section). We show in particular that our method outperforms SRGAN on the satellite images domain.
\subsubsection{Datasets}
\label{sec:Datasets}

	The idea of replacing the MSE pixel-wise content loss on the image by a loss function that is closer to perceptual similarity is not new. Indeed, \cite{ledig2016photo} defined a VGG loss on the feature map obtained by a specific layer of the pre-trained VGG19 network and shows that it fixes the inherent problem of overly smooth results which comes with the pixel-wise loss. Nevertheless, VGG19 being trained on ImageNet, their method would not perform particularly well on different images, the distribution of which is far away from that of ImageNet.
Therefore, we propose a similar method where the difference is that our features extractor is not pre-trained, but trained jointly with the generator. This removes the aforementioned limitation since the features extractor is trained on the same dataset as the generator and thus extract relevant features. 
 
\begin{figure}[t!]
\centering
\includegraphics[width= \linewidth]{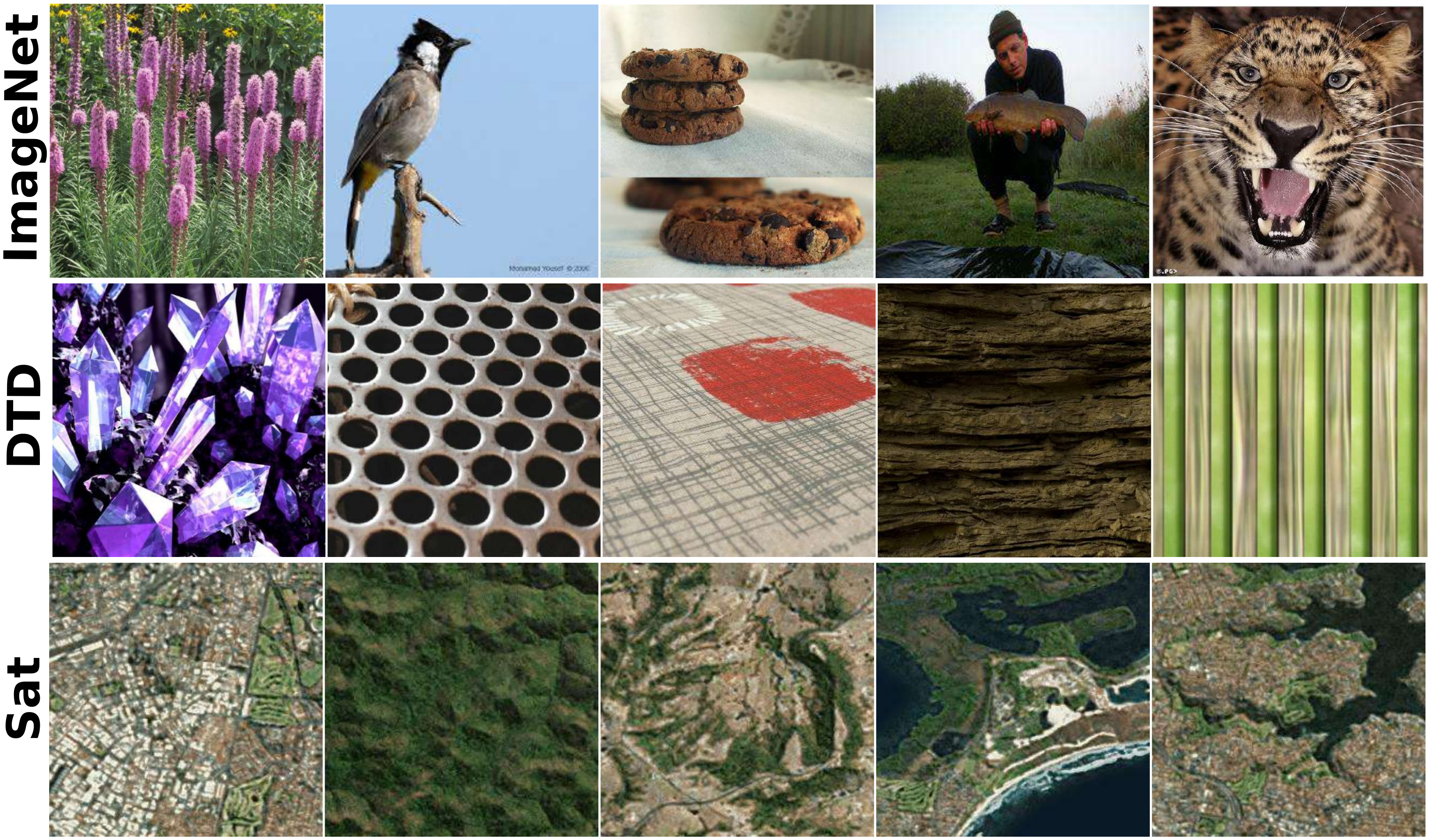}
\caption{Examples of images from the considered datasets.}
\label{fig:datasets}
\end{figure}

	To show that, we trained our different networks (\textit{i.e.,} with different features extractors) on three distinct datasets (examples of images of these datasets are shown in Figure~\ref{fig:datasets}):
    \begin{itemize}
    \item A subset of \textit{ImageNet} \cite{deng2009imagenet}, for which we sampled $70,000$ images. Since VGG19 was trained on ImageNet for many (more than 300K) iterations, we expect to have similar or worse results than the state-of-the-art method SRGAN from \cite{ledig2016photo} on this database.
    \item \textit{The Describable Textures Dataset (DTD)} \cite{cimpoi14describing}, containing $5,600$ images of textural patterns. These data are relatively close to ImageNet and we show that our method gives convincing results relatively close to SRGAN.
    \item A dataset containing satellite images\footnote{Can be found in \url{http://www.terracolor.net/sample_imagery.html}}, which we generated by randomly cropping $256\times 256$ images on a $7205\times 7205$ satellite image which result in $235,183$ images. We particularly show that our method significantly outperforms SRGAN on this dataset. We refer to this dataset by \textit{Sat}.
    \end{itemize}
 
All experiments are performed with a scale factor of $4\times$ between low- and high-resolutions images and the formers are obtained during the training by down-scaling the original images by a factor $1/4$.

\subsubsection{Architectures}
\begin{figure}[t!]
	    \centering
	    \includegraphics[width=\textwidth]{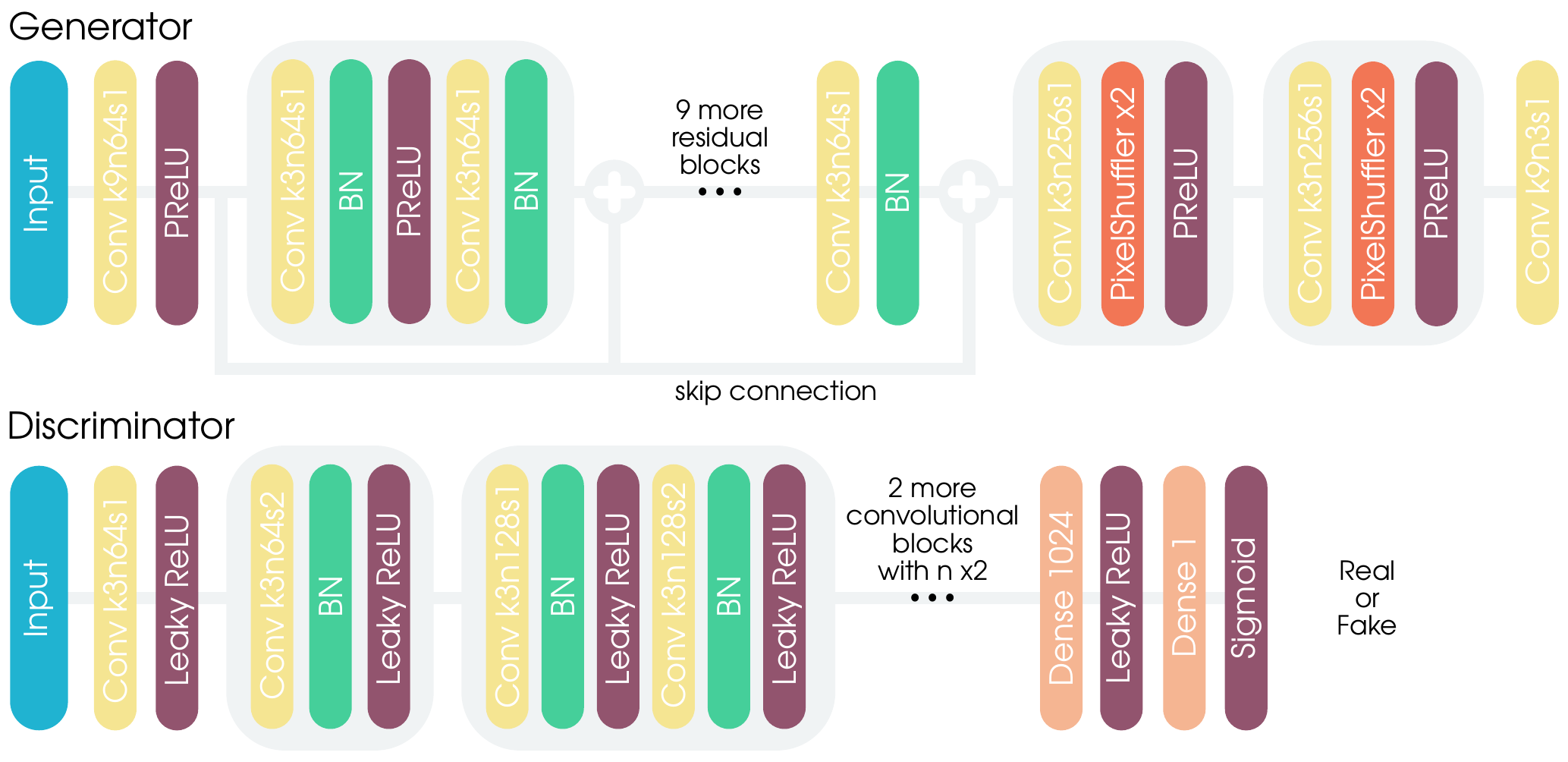}
	    \caption{Overview of the used architectures for the generator and the discriminator. We have considered the same architectures as that of Ledig \textit{et al.}~\cite{ledig2016photo}.}
	    \label{fig:architecture}
	\end{figure}

	Our overall goal is to prove that the proposed GCN framework, is adapted to train a generative mapping model and that it surpasses the MSE loss in keeping perceptual similarity in the generated image (whereas the MSE loss tends to smooth things out and lose high frequency details). As opposed to \cite{ledig2016photo}'s work, our framework does not require to have a pre-trained network, like VGG, to extract helpful features for training. In this paper, we focus on the Super Resolution problem. Therefore, we chose our mapping function $\mathcal{F}$, or generator, to be that of Ledig \textit{et al.} \cite{ledig2016photo}: a feed-forward CNN parametrized by $\theta_{\mathcal{F}}$, composed of $10$ residual blocks. These blocks are made of two convolutional layers with $3\times 3$ kernels and $64$ features maps, each followed by batch normalization and PReLU as activation. The image's size is then increased of a factor $4$ by two trained upsamplings. The architecture of all the used discriminators follows the guidelines of Radford \textit{et al.} \cite{radford2015unsupervised} as it is composed of convolutional layers, followed by a batch normalization and a LeakyReLU ($\alpha = 0.2$) activation. This block is repeated eight times and each time the number of $3\times 3$ kernels increases by a factor $2$ (ranging from $64$ to $512$), a strided convolution is used to reduce the image resolution by $2$. Two dense layers and a sigmoid activation then return the discrimination probability. In the case of an auto-encoder (every $\textit{Reconstruction}$ problem), we follow the same architecture for the encoder and a symmetric one for the decoder. Figure~\ref{fig:architecture} depicts an overview of the architectures for both the generator and the discriminator.
	
\subsubsection{Training details and parameters}
	All networks were trained\footnote{A Keras implementation is provided in \url{https://github.com/melaseddik/GCN}} on a NVIDIA Geoforce GTX 1070 GPU using the datasets described in Section~\ref{sec:Datasets}, which do not contain the $(1000)$ testing images shown as results.  We scaled the range of both the LR input images and the HR images to $[-1, 1]$, which explains the $\tanh$ activation for the last layer of the generator. All variants of our networks, which differ in their features extractor, were trained from scratch (for the generator and the features extractor) with mini batches of 10 images. We used the Adam optimizer with a learning rate of $2\cdot 10^{-4}$ and a decay of $0$. The generator and the feature extractor are updated alternatively. As we realized training was stable and quite fast, we trained with only $5,000$ update iterations to pinpoint the best method among the different GCNs. Finally, the regularization parameters in our global loss are set by default as $\lambda_1=1$ and $\lambda=10^{-3}.$ As a reminder, our goal here is, given a generator architecture (or mapping function $\mathcal{F}$), to find the best strategy to train it, following our GCNs paradigms. The best method is then further compared to baselines.

\subsubsection{Features Extractor Selection}

	As we said above, we investigated the ability of different features extractor to construct relevant perceptual feature maps for training and improving the rendering quality of the generator. In order to select the best learning strategy given a certain dataset, we train the generator on each dataset (presented in Section~\ref{sec:Datasets}) using the different learning strategies: $\mathcal{P}_{sisr}$/\textit{rec}, $\mathcal{P}_{sisr}$/\textit{dis}, $\mathcal{P}_{sisr}$/\textit{dis,rec}, $\mathcal{P}_{sisr}$/\textit{adv} and $\mathcal{P}_{sisr}$/\textit{adv,rec}. Note that, the features extractor for all the considered methods correspond to the first layer of the discriminators (or encoder-decoders). In fact, as the problem $\mathcal{P}_{sisr}$ consists in recovering low-level perceptual cues, we limited our study to the first layer.  

	Table~\ref{tab:ours} summarizes the results of the proposed $\mathcal{P}_{sisr}$ methods in terms of low-level metrics (L2 and SSIM) and perceptual metrics \citep{zhang2018unreasonable} which are given by Eq.~\eqref{eq:PE_metric}. We notice from this table that the method $\mathcal{P}_{sisr}/adv,rec$ performs relatively well on the datasets ImageNet and Sat in terms of perceptual metrics. While $\mathcal{P}_{sisr}/dis,rec$ gives better results on the DTD dataset. The main difference between these two methods is that the former considers an adversarial loss on the objective function while the latter does not consider the adversarial term. This explains the reason why $\mathcal{P}_{sisr}/adv,rec$ does not perform well on DTD. In fact, texture images belong to a complex manifold and their distribution is relatively hard to fit by a generative model. 
    
\begin{figure}
  \begin{adjustbox}{addcode={\begin{minipage}{\width}}{\caption{%
      Rows refer to the different considered Datasets. Columns refer to methods and ground-truth images: \textbf{LR} and \textbf{HR} refer to the low- and high-resolution pairs. The different used methods are: \textbf{(a)} $\mathcal{P}_{sisr}$/\textit{rec}, \textbf{(b)} $\mathcal{P}_{sisr}$/\textit{dis,rec}, \textbf{(c)} $\mathcal{P}_{sisr}$/\textit{adv} and \textbf{(d)} $\mathcal{P}_{sisr}$/\textit{adv,rec}. Best view in PDF.\label{fig:ours}
      }\end{minipage}},rotate=90,center}
      \includegraphics[height=\linewidth]{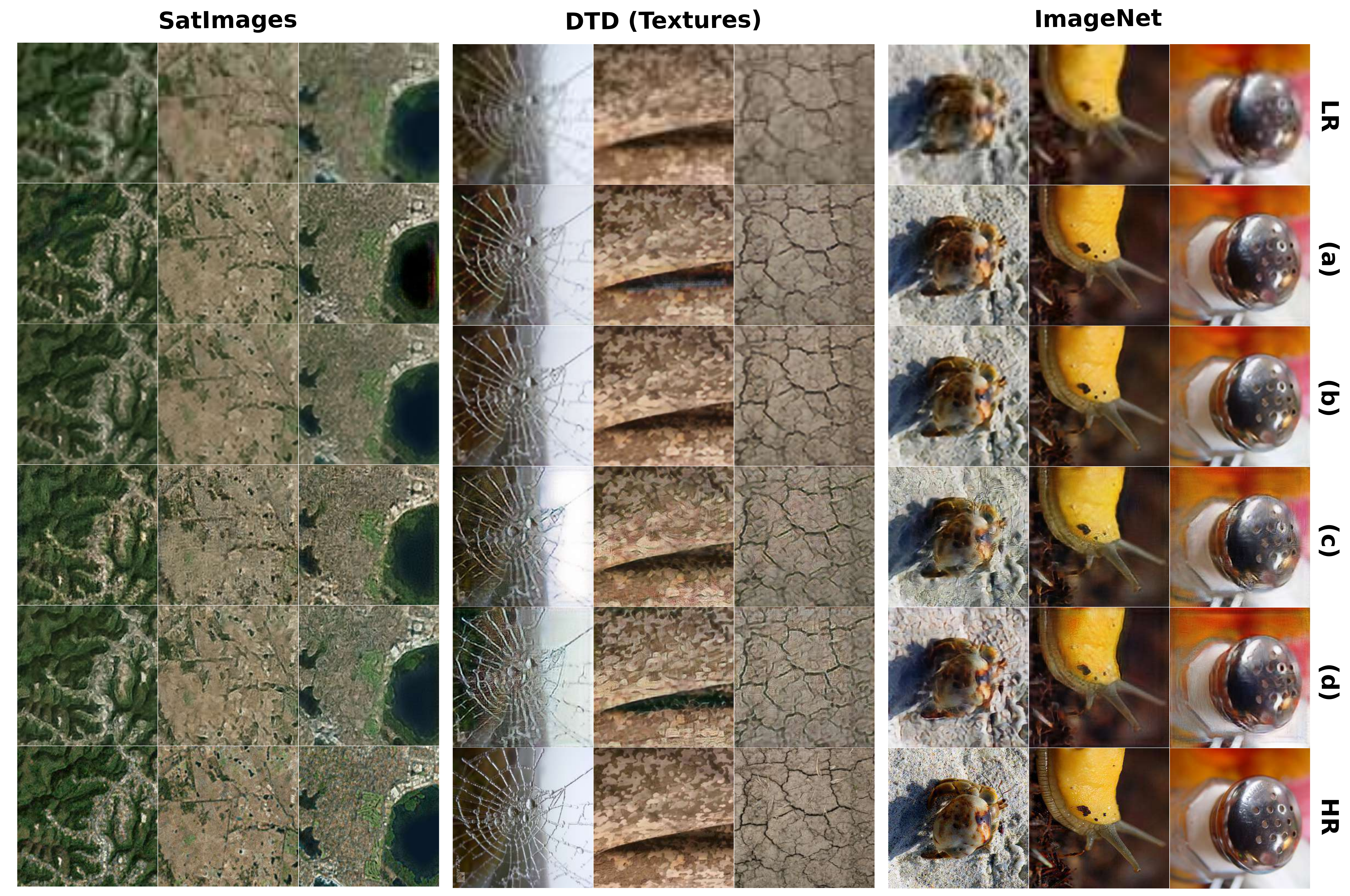}%
  \end{adjustbox}
\end{figure}

    Figure~\ref{fig:ours} shows qualitative results of the different proposed methods on the different presented datasets. Generally, the methods which were trained with an additional adversarial loss ($\mathcal{P}_{sisr}$/\textit{adv} and $\mathcal{P}_{sisr}$/\textit{adv,rec}) output images of higher quality (on the datasets ImageNet and Sat) as GANs were introduced to do just so: generate images that follow the distribution of the dataset. Among these two \textit{adversarial} methods, it seems to us (as suggested by the quantitative results of table~\ref{tab:ours}) that $\mathcal{P}_{sisr}$/\textit{adv,rec} (column (c) of Figure \ref{fig:ours}) is able to detect and render more details, due to its ability to generate more relevant features as the features extractor $\Phi$ is learned to solve a \textit{multi-task} problem; namely a \textit{discrimination} and a \textit{reconstruction} problem, in particular, this method allows for the learning of both classification and reconstruction-based features. We will thus further investigate the $\mathcal{P}_{sisr}$/\textit{adv,rec} method for the comparison to the baseline and the state-of-the-art method SRGAN \citep{ledig2016photo}, on the satellite images domain.

\begin{table}[H]
\scriptsize
\centering
\begin{tabular}{| c | l | c c | c c c c c c |}
\multicolumn{2}{c}{} & \multicolumn{2}{c}{Low-level} & \multicolumn{6}{c}{Perceptual metrics} \\
\hline
  & \textbf{Methods} & L2 & SSIM & Squ & Squ-l & Alex & Alex-l & VGG & VGG-l\\ 
\hline
\hline
\parbox[t]{1mm}{\multirow{5}{*}{\rotatebox[origin=c]{90}{ImageNet}}}
& $\mathcal{P}_{sisr}/dis$     & 0.018 & \underline{0.147} & 1.606 & 0.279 & 1.470 & 0.398 & 2.088 & 0.358\\
& $\mathcal{P}_{sisr}/rec$     & 0.020 & 0.162 & 1.723 & 0.301 & 1.595 & 0.425 & 2.243 & 0.388\\
& $\mathcal{P}_{sisr}/dis,rec$ & \underline{0.017} & \underline{0.147} & \underline{1.587} & 0.279 & \underline{1.420} & 0.382 & \underline{2.052} & \underline{0.353}\\
& $\mathcal{P}_{sisr}/adv$     & 0.028 & 0.202 & 1.820 & \textbf{0.222} & 1.554 & \textbf{0.322} & 2.598 & 0.432\\
& $\mathcal{P}_{sisr}/adv,rec$ & \textbf{0.016} & \textbf{0.141} & \textbf{1.533} & \underline{0.263} & \textbf{1.362} & \underline{0.368} & \textbf{1.994} & \textbf{0.340}\\
\hline
\hline
\parbox[t]{1mm}{\multirow{5}{*}{\rotatebox[origin=c]{90}{DTD}}}
& $\mathcal{P}_{sisr}/dis$     & \underline{0.027} & 0.184 & 1.873 & 0.327 & 1.739 & 0.440 & 2.401 & 0.421\\
& $\mathcal{P}_{sisr}/rec$     & \underline{0.027} & \underline{0.183} & \underline{1.851} & 0.320 & \underline{1.726} & 0.438 & \underline{2.398} & \underline{0.420}\\
& $\mathcal{P}_{sisr}/dis,rec$ & \textbf{0.023} & \textbf{0.167} & \textbf{1.703} & 0.292 & \textbf{1.576} & 0.404 & \textbf{2.260} & \textbf{0.392}\\
& $\mathcal{P}_{sisr}/adv$     & 0.036 & 0.227 & 2.077 & \underline{0.281} & 1.812 & \underline{0.375} & 2.770 & 0.473 \\
& $\mathcal{P}_{sisr}/adv,rec$ & 0.046 & 0.236 & 2.089 & \textbf{0.277} & 1.793 & \textbf{0.344} & 2.796 & 0.481\\
\hline
\hline
\parbox[t]{1mm}{\multirow{5}{*}{\rotatebox[origin=c]{90}{Sat}}}
& $\mathcal{P}_{sisr}/dis$     & \textbf{0.011} & \textbf{0.129} & \underline{1.484} & 0.210 & 1.508 & 0.356 & 2.121 & \underline{0.355}\\
& $\mathcal{P}_{sisr}/rec$     & 0.060 & 0.168 & 1.705 & 0.245 & 1.762 & 0.423 & 2.260 & 0.395\\
& $\mathcal{P}_{sisr}/dis,rec$ & \textbf{0.011} & \underline{0.138} & 1.493 & 0.215 & \underline{1.435} & 0.351 & \textbf{2.108} & 0.372\\
& $\mathcal{P}_{sisr}/adv$     & 0.030 & 0.214 & 1.719 & \underline{0.181} & 1.627 & \underline{0.306} & 2.711 & 0.419\\
& $\mathcal{P}_{sisr}/adv,rec$ & \underline{0.018} & 0.183 & \textbf{1.359} & \textbf{0.140} & \textbf{1.310} & \textbf{0.220} & \underline{2.115} & \textbf{0.344}\\
\hline
\end{tabular}
\caption{Results of the proposed $\mathcal{P}_{sisr}$ methods in terms of traditional metrics (L2 and SSIM) and the \textit{perceptual error} (PE) given by Eq.~\eqref{eq:PE_metric} on different datasets. As we can notice, the method $\mathcal{P}_{sisr}/adv,rec$ outperforms the other methods in the datasets ImageNet and Sat, while $\mathcal{P}_{sisr}/dis,rec$ gives the best results on DTD.}
\label{tab:ours}
\end{table}

\subsubsection{$\mathcal{P}_{sisr}/adv,rec$ against baseline methods on the satellite images domain}
	Our main objective is to show that the VGG loss function (namely, the SRGAN method \citep{ledig2016photo}) is no longer relevant when super-resolving images from a domain different than the ImageNet domain. In particular, by considering the satellite images domain, we show in this section that the selected method from the previous section ($\mathcal{P}_{sisr}/adv,rec$) outperforms some baselines, which are $\mathcal{P}_{sisr}$/\textit{mse} (pixel-wise MSE loss) and $\mathcal{P}_{sisr}$/\textit{adv,mse} (pixel wise MSE loss combined with an adversarial loss), and the state-of-the-art super-resolution method, SRGAN \citep{ledig2016photo}. Note that all the methods use the same architectures (depicted in figure~\ref{fig:architecture}) for the generator and discriminator and are trained on the same domain (here, on satellite images). Our purpose being to show the relevance of the proposed method on a domain ``far'' from the ImageNet domain, we do not consider standard SR benchmarks, which are raltively ``close'' to the ImageNet domain.
    
    Table~\ref{tab:res_sat} presents quantitative results, in terms of classical metrics (L2 and SSIM) and perceptual metrics given by Eq.~\eqref{eq:PE_metric}, of the different methods on the Sat dataset. As we can notice, our method $\mathcal{P}_{sisr}/adv,rec$ outperforms the other methods in terms of perceptual metrics. Knowing that the perceptual metrics agree with human judgments \citep{zhang2018unreasonable}, these results validate the effectiveness of the $\mathcal{P}_{sisr}/adv,rec$ method. Note also that even if SRGAN \citep{ledig2016photo} is optimized to minimize a VGG loss, it does not give the lowest perceptual errors in terms of the perceptual metrics VGG and VGG-l, this is due to the fact that the VGG features are not relevant for the satellite images domain. In addition, $\mathcal{P}_{sisr}/adv,rec$ gives the lowest perceptual errors in terms of the perceptual metrics Alex and Alex-l which agrees with a human perception. In fact, AlexNet network may more closely match the architecture of the human visual cortex \citep{yamins2016using}.

\begin{table}[t!]
\scriptsize
\centering
\begin{tabular}{| c | l | c c | c c c c c c |}
\multicolumn{2}{c}{} & \multicolumn{2}{c}{Low-level} & \multicolumn{6}{c}{Perceptual metrics} \\
\hline
  & \textbf{Methods} & L2 & SSIM & Squ & Squ-l & Alex & Alex-l & VGG & VGG-l\\ 
\hline
\hline
\parbox[t]{1mm}{\multirow{4}{*}{\rotatebox[origin=c]{90}{Sat}}}
& $\mathcal{P}_{sisr}/mse$     & \textbf{0.011} & \textbf{0.134} & 1.873 & 0.245 & 1.855 & 0.411 & 2.536 & 0.419\\
& $\mathcal{P}_{sisr}/adv,mse$ & 0.082 & 0.197 & \underline{1.458} & \underline{0.205} & 1.466 & 0.352 & \underline{2.125} & \underline{0.347}\\
& SRGAN \citep{ledig2016photo}                       & 0.228 & 0.188 & 1.510 & 0.220 & \underline{1.361} & \underline{0.282} & 2.230 & 0.412\\
& $\mathcal{P}_{sisr}/adv,rec$ & \underline{0.018} & \underline{0.183} & \textbf{1.359} & \textbf{0.140} & \textbf{1.310} & \textbf{0.220} & \textbf{2.115} & \textbf{0.344}\\
\hline
\end{tabular}
\caption{Comparison of our method $\mathcal{P}_{sisr}/adv,rec$ with baselines and the SRGAN method \citep{ledig2016photo} on the satellite images domain, in terms of classical metrics (L2 and SSIM) and perceptual metrics \citep{zhang2018unreasonable}.}
\label{tab:res_sat}
\end{table}

\captionsetup[subfigure]{labelformat=empty}
   \begin{figure}[b!]
   \centering
     \subfloat[]{%
       \includegraphics[width=0.3\textwidth]{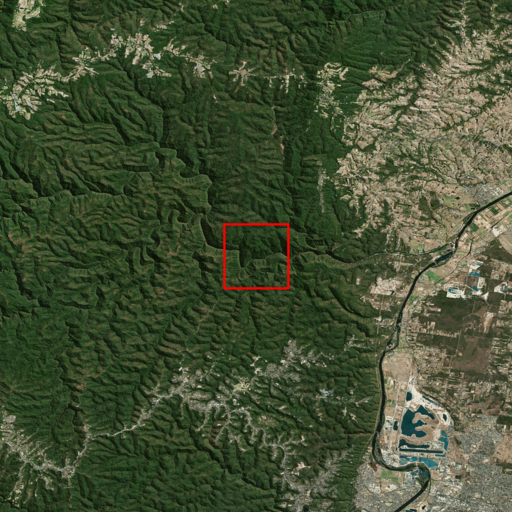}
     }
     \hfill
     \raisebox{2.1cm}[0pt][0pt]{\subfloat[HR (REF)]{%
       \includegraphics[width=0.15\textwidth]{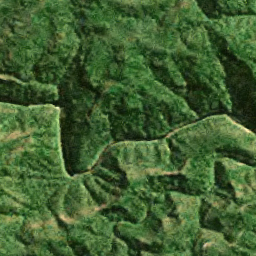}
     }}
     \raisebox{2.1cm}[0pt][0pt]{\subfloat[$\mathcal{P}_{sisr}$/\textit{mse}]{%
       \includegraphics[width=0.15\textwidth]{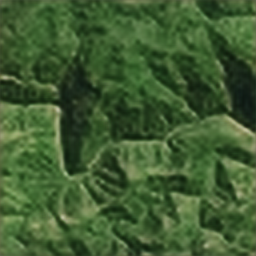}
     }}
     \raisebox{2.1cm}[0pt][0pt]{\subfloat[$\mathcal{P}_{sisr}$/\textit{adv,mse}]{%
       \includegraphics[width=0.15\textwidth]{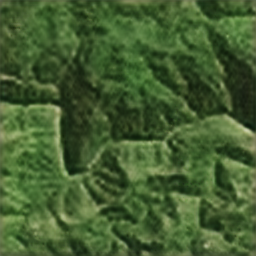}
     }}
     \raisebox{2.1cm}[0pt][0pt]{\subfloat[SRGAN \citep{ledig2016photo}]{%
       \includegraphics[width=0.15\textwidth]{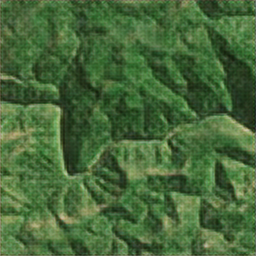}
     }}
     \hspace*{3.92cm} 
     \raisebox{0.1cm}[0pt][0pt]{\subfloat[$\mathcal{P}_{sisr}$/\textit{rec}]{%
       \includegraphics[width=0.15\textwidth]{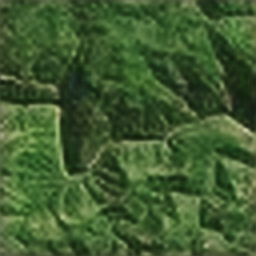}
     }}
     \raisebox{0.1cm}[0pt][0pt]{\subfloat[$\mathcal{P}_{sisr}$/\textit{dis,rec}]{%
       \includegraphics[width=0.15\textwidth]{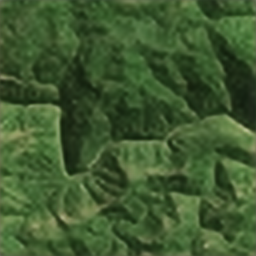}
     }}
     \raisebox{0.1cm}[0pt][0pt]{\subfloat[$\mathcal{P}_{sisr}$/\textit{adv}]{%
       \includegraphics[width=0.15\textwidth]{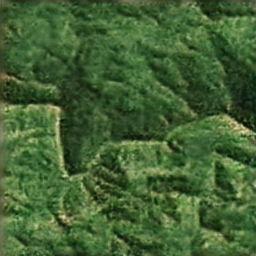}
     }}
     \raisebox{0.1cm}[0pt][0pt]{\subfloat[$\mathcal{P}_{sisr}$/\textit{adv,rec}]{%
       \includegraphics[width=0.15\textwidth]{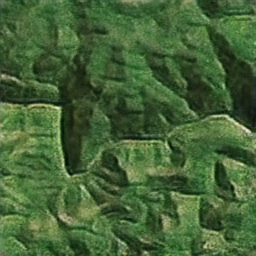}
     }}
     \vspace{-.5cm}
     \caption{Results of different $\mathcal{P}_{sisr}$ methods on a patch of an image from the Sat dataset.}
     \label{fig:res_sat}
   \end{figure}

	Figure~\ref{fig:res_sat} shows some qualitative results of different methods on a patch of an image from the Sat dataset. As we can notice, the $\mathcal{P}_{sisr}/adv,rec$ method gives the perceptually closest result to the ground-truth image, which agrees with the quantitative results of table~\ref{tab:res_sat}.

\subsubsection{Further results}
\begin{figure}[H]
  \begin{adjustbox}{addcode={\begin{minipage}{\width}}{\caption{%
      Rows refer to the different Datasets. Columns refer to methods and ground-truth images: \textbf{LR} and \textbf{HR} refer to the low- and high-resolution pairs. \textbf{P-mse+} refers to the method $\mathcal{P}_{sisr}$/\textit{mse} with an adversarial loss ($\lambda_2>0$), \textbf{SRGAN} for the method in \citep{ledig2016photo} and our method $\mathcal{P}_{sisr}$/\textit{adv,rec}. Best view in PDF.\label{fig:srgan}
      }\end{minipage}},rotate=90,center}
      \includegraphics[height=\linewidth]{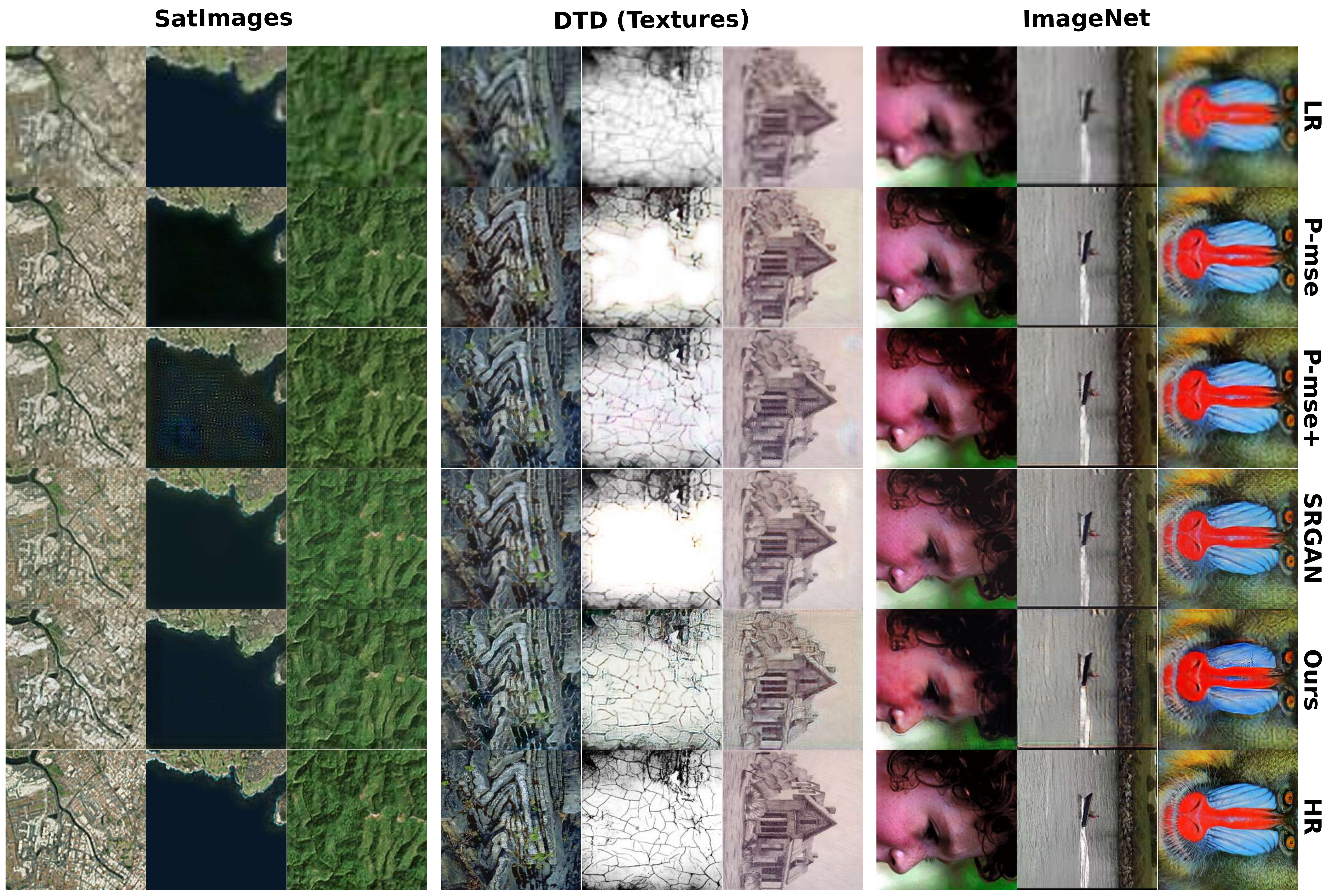}%
  \end{adjustbox}
\end{figure}

	In this section, we provide further qualitative and quantitative comparisons to the considered baselines of the previous section. In particular, we consider all the presented datasets for the comparisons. Qualitative results are provided in figure \ref{fig:srgan}. SRGAN performs better on ImageNet, which is not that surprising considering our features extractor was trained much less than VGG19 used in \cite{ledig2016photo} and the VGG features being more relevant for images from the ImageNet domain. Nonetheless, we do have sharper images than the MSE based methods, although we show some artifact (especially on the boat) which we attribute to the competition between the content and adversarial losses. On DTD though, we can see the benefit of our method over a pre-trained VGG loss. Indeed, SRGAN is blurrier on both the house (first row) and the cliff (third row), in spite of having less artifacts than our method. On the ``cracks'' example (second row), SRGAN even totally obliterates the details in the center. Finally, results on the dataset Sat, which is the most different dataset compared to ImageNet, are the most compelling. Our method generates super resolved images that are really close to the real high resolution images, while we can clearly see imperfections on SRGAN's results because of VGG19 which was not trained to detect perceptual features on satellite images.

\begin{table}[t!]
\scriptsize
\centering
\begin{tabular}{| c | l | c c | c c c c c c |}
\multicolumn{2}{c}{} & \multicolumn{2}{c}{Low-level} & \multicolumn{6}{c}{Perceptual metrics} \\
\hline
  & \textbf{Methods} & L2 & SSIM & Squ & Squ-l & Alex & Alex-l & VGG & VGG-l\\ 
\hline
\hline
\parbox[t]{1mm}{\multirow{4}{*}{\rotatebox[origin=c]{90}{ImageNet}}}
& $\mathcal{P}_{sisr}/mse$     & \underline{0.017} & \underline{0.146} & 1.568 & 0.280 & 1.435 & 0.391 & 2.064 & 0.349\\
& $\mathcal{P}_{sisr}/adv,mse$ & 0.020 & 0.156 & 1.634 & \underline{0.241} & 1.397 & \underline{0.329} & 2.223 & 0.384\\
&  SRGAN                        & 0.028 & 0.170 & \textbf{1.303} & \textbf{0.177} & \textbf{1.084} & \textbf{0.225} & \underline{2.045} & \underline{0.342}\\
& $\mathcal{P}_{sisr}/adv,rec$ & \textbf{0.016} & \textbf{0.141} & \underline{1.533} & 0.263 & \underline{1.362} & 0.368 & \textbf{1.994} & \textbf{0.340}\\
\hline
\hline
\parbox[t]{1mm}{\multirow{4}{*}{\rotatebox[origin=c]{90}{DTD}}}
& $\mathcal{P}_{sisr}/mse$     & \underline{0.029} & \underline{0.185} & 1.972 & 0.342 & 1.856 & 0.470 & 2.479 & 0.434\\
& $\mathcal{P}_{sisr}/adv,mse$ & 0.025 & 0.188 & 1.880 & \underline{0.268} & 1.586 & 0.349 & 2.512 & 0.430\\
& SRGAN                        & 0.031 & 0.191 & \textbf{1.557} & \textbf{0.209} & \textbf{1.298} & \textbf{0.241} & \underline{2.308} & \underline{0.393}\\
& $\mathcal{P}_{sisr}/dis,rec$ & \textbf{0.023} & \textbf{0.167} & \underline{1.703} & 0.292 & \underline{1.576} & 0.404 & \textbf{2.260} & \textbf{0.392}\\
\hline
\end{tabular}
\caption{Comparison of our methods $\mathcal{P}_{sisr}/adv,rec$ and $\mathcal{P}_{sisr}/dis,rec$ with baselines and the SRGAN method \citep{ledig2016photo} on the datasets ImageNet (a subset of 200,000 randomely selected images) and DTD, in terms of classical metrics (L2 and SSIM) and perceptual metrics \citep{zhang2018unreasonable}.}
\label{tab:all}
\end{table}
Quantitative results are summarized in Table~\ref{tab:all}. As shown in \citep{ledig2016photo,zhang2018unreasonable}, the standard quantitative measures such as L2 and SSIM fail to highlight image quality according to the human visual system. In fact, while the results of $\mathcal{P}_{sisr}$/\textit{mse} are overly smooth perceptually, it has the lowest L2 and SSIM errors on Sat. However, perceptual metrics agree with what we assess qualitatively: SRGAN performs best on ImageNet but not on Sat, the distribution of which is the farthest from ImageNet. Actually, SRGAN ranks third of all four methods on Sat, just before $\mathcal{P}_{sisr}$/\textit{adv,mse}, while still performing best on DTD which still is pretty close to ImageNet. This shows that the VGG features become less and less relevant as the dataset's distribution part from ImageNet. On the other hand, our training framework allows to construct relevant features on any (never seen) dataset. Thus our method $\mathcal{P}_{sisr}$/\textit{adv,rec} performs best on Sat. Our method performing better than $\mathcal{P}_{sisr}$/\textit{adv,mse} also shows that our framework helps finding detail preserving features. Figure~\ref{fig:srgan} provides the results of the different baselines and our method on some examples of the considered datasets. We notice from these images that our method $\mathcal{P}_{sisr}/adv,rec$ recovers finer details on the different datasets while it outperforms the considered baselines on satellite images.
Table~\ref{tab:all_methods} summarizes the results of the different methods on the considered datasets through the paper. From these results, we make the following conclusions:
\begin{itemize}
\item When the considered domain is far enough from the ImageNet domain, the VGG loss introduced by \citep{ledig2016photo} is no longer relevant.
\item The VGG network can not be fine-tuned when considering a domain for which there is no available labels for the images (\textit{e.g.}, satellite images). Thus, the SRGAN method cannot be exploited efficiently in this case.
\item Our framework results in a method ($\mathcal{P}_{sisr}/adv,rec$) that outperforms some baselines and the SRGAN method on the satellite images domain.
\item Even on a domain close to the ImageNet domain (\textit{e.g.}, texture images), one can find within our framework methods which give almost similar results to the SRGAN method, while the later is based on VGG features and thus need to train the VGG network on the whole ImageNet dataset. 
\end{itemize}

\begin{table}[H]
\scriptsize
\centering
\begin{tabular}{| c | l | c c | c c c c c c |}
\multicolumn{2}{c}{} & \multicolumn{2}{c}{Low-level} & \multicolumn{6}{c}{Perceptual metrics} \\
\hline
  & \textbf{Methods} & L2 & SSIM & Squ & Squ-l & Alex & Alex-l & VGG & VGG-l\\ 
\hline
\hline
\parbox[t]{1mm}{\multirow{8}{*}{\rotatebox[origin=c]{90}{ImageNet}}}
& $\mathcal{P}_{sisr}/mse$     & \underline{0.017} & \underline{0.146} & 1.568 & 0.280 & 1.435 & 0.391 & 2.064 & 0.349\\
& $\mathcal{P}_{sisr}/adv,mse$ & 0.020 & 0.156 & 1.634 & 0.241 & 1.397 & 0.329 & 2.223 & 0.384\\
&  SRGAN                        & 0.028 & 0.170 & \textbf{1.303} & \textbf{0.177} & \textbf{1.084} & \textbf{0.225} & \underline{2.045} & \underline{0.342}\\
& $\mathcal{P}_{sisr}/dis$     & 0.018 & 0.147 & 1.606 & 0.279 & 1.470 & 0.398 & 2.088 & 0.358\\
& $\mathcal{P}_{sisr}/rec$     & 0.020 & 0.162 & 1.723 & 0.301 & 1.595 & 0.425 & 2.243 & 0.388\\
& $\mathcal{P}_{sisr}/dis,rec$ & \underline{0.017} & 0.147 & 1.587 & 0.279 & 1.420 & 0.382 & 2.052 & 0.353\\
& $\mathcal{P}_{sisr}/adv$     & 0.028 & 0.202 & 1.820 & \underline{0.222} & 1.554 & \underline{0.322} & 2.598 & 0.432\\
& $\mathcal{P}_{sisr}/adv,rec$ & \textbf{0.016} & \textbf{0.141} & \underline{1.533} & 0.263 & \underline{1.362} & 0.368 & \textbf{1.994} & \textbf{0.340}\\
\hline
\hline
\parbox[t]{1mm}{\multirow{8}{*}{\rotatebox[origin=c]{90}{DTD}}}
& $\mathcal{P}_{sisr}/mse$     & 0.029 & 0.185 & 1.972 & 0.342 & 1.856 & 0.470 & 2.479 & 0.434\\
& $\mathcal{P}_{sisr}/adv,mse$ & 0.025 & 0.188 & 1.880 & \underline{0.268} & 1.586 & 0.349 & 2.512 & 0.430\\
& SRGAN                        & 0.031 & 0.191 & \textbf{1.557} & \textbf{0.209} & \textbf{1.298} & \textbf{0.241} & \underline{2.308} & \underline{0.393}\\
& $\mathcal{P}_{sisr}/dis$     & \underline{0.027} & 0.184 & 1.873 & 0.327 & 1.739 & 0.440 & 2.401 & 0.421\\
& $\mathcal{P}_{sisr}/rec$     & \underline{0.027} & \underline{0.183} & 1.851 & 0.320 & 1.726 & 0.438 & 2.398 & 0.420\\
& $\mathcal{P}_{sisr}/dis,rec$ & \textbf{0.023} & \textbf{0.167} & \underline{1.703} & 0.292 & \underline{1.576} & 0.404 & \textbf{2.260} & \textbf{0.392}\\
& $\mathcal{P}_{sisr}/adv$     & 0.036 & 0.227 & 2.077 & 0.281 & 1.812 & 0.375 & 2.770 & 0.473 \\
& $\mathcal{P}_{sisr}/adv,rec$ & 0.046 & 0.236 & 2.089 & 0.277 & 1.793 & \underline{0.344} & 2.796 & 0.481\\
\hline
\hline
\parbox[t]{1mm}{\multirow{8}{*}{\rotatebox[origin=c]{90}{Sat}}}
& $\mathcal{P}_{sisr}/mse$     & \textbf{0.011} & \underline{0.134} & 1.873 & 0.245 & 1.855 & 0.411 & 2.536 & 0.419\\
& $\mathcal{P}_{sisr}/adv,mse$ & 0.082 & 0.197 & \underline{1.458} & 0.205 & 1.466 & 0.352 & 2.125 & \underline{0.347}\\
& SRGAN                        & 0.228 & 0.188 & 1.510 & 0.220 & 1.361 & 0.282 & 2.230 & 0.412\\
& $\mathcal{P}_{sisr}/dis$     & \textbf{0.011} & \textbf{0.129} & 1.484 & 0.210 & 1.508 & 0.356 & 2.121 & 0.355\\
& $\mathcal{P}_{sisr}/rec$     & 0.060 & 0.168 & 1.705 & 0.245 & 1.762 & 0.423 & 2.260 & 0.395\\
& $\mathcal{P}_{sisr}/dis,rec$ & \textbf{0.011} & 0.138 & 1.493 & 0.215 & \underline{1.435} & 0.351 & \textbf{2.108} & 0.372\\
& $\mathcal{P}_{sisr}/adv$     & 0.030 & 0.214 & 1.719 & \underline{0.181} & 1.627 & \underline{0.306} & 2.711 & 0.419\\
& $\mathcal{P}_{sisr}/adv,rec$ & \underline{0.018} & 0.183 & \textbf{1.359} & \textbf{0.140} & \textbf{1.310} & \textbf{0.220} & \underline{2.115} & \textbf{0.344}\\
\hline
\end{tabular}
\caption{Comparison of the proposed $\mathcal{P}_{sisr}$ methods in terms of traditional metrics (L2 and SSIM) and the \textit{perceptual error} (PE) given by Eq.~\eqref{eq:PE_metric} on all the considered datasets. In terms of perceptual metrics, the proposed $\mathcal{P}_{sisr}$ methods rank in the second position after SRGAN \citep{ledig2016photo} on the datasets ImageNet and DTD, while they outperform all the baselines on the satellite images domain which is far from the ImageNet domain.}
\label{tab:all_methods}
\end{table}

\section{Conclusion and Perspectives}
\label{sec:disc}
In this paper, we propose a general framework named Generative Collaborative Networks (GCN) which generalizes the existing methods for the problem of learning a mapping between two domains. The GCN framework highlights that there is a learning strategy of mappings that is not explored in the literature. In particular, the optimization of these mappings in the feature space of a features extractor network, which is mutually learned at the same time as the considered mapping (\textit{joint-learning} strategy). The GCN framework was evaluated in the context of super-resolution on three datasets (ImageNet \cite{deng2009imagenet}, \textit{DTD} \cite{cimpoi14describing} and satellite images). We have shown that the proposed \textit{joint-learning} strategy leads to a method that outperforms the state of the art \cite{ledig2016photo} which uses a pre-trained features extractor network (VGG19 on ImageNet). Specifically, this holds when the domain of interest is ``far'' from the ImageNet domain (\textit{e.g.,} satellite images or images from the medical domain\footnote{This domain is particularly relevant for the proposed framework as it seems very far from the ImageNet domain. Unfortunately, we have not found a big amount of publicly available data (to the best of our knowledge) for medical images which prevented us from considering this domain through the paper.}). However, note that even for domains close to the ImageNet domain, the proposed method gives convincing (almost similar to \cite{ledig2016photo}) results without using the whole ImageNet dataset to learn the features extractor network (as performed in \cite{ledig2016photo}).

In this work, we systematically designed the proposed methods by using the first layer of the features extractor networks, while it could be interesting to evaluate in more detail the impact of this choice regarding the learning strategy. Moreover, the impact of the selected layer may also depend on the considered dataset. More generally, the GCN framework offers a large vision on the wide variety of existing loss functions used in the literature of learning mappings-based problems (\textit{e.g.,} super-resolution, image completion, artistic style transfer, etc.). In fact, we show that these loss functions can be simply reformulated, in the proposed framework, as a certain combination of a particular type of features extractor networks ($\mathcal{P}$/\textit{rec}, $\mathcal{P}$/\textit{dis}, $\mathcal{P}$/\textit{dis,rec}, $\mathcal{P}$/\textit{adv} and $\mathcal{P}$/\textit{adv,rec}) and a particular learning strategies (\textit{joint-learning} or \textit{disjoint-learning}). Therefore it will be interesting to explore this promising framework in other learning mappings-based problems.


\bibliography{mybibfile}

\end{document}